\RequirePackage{fix-cm}
\documentclass[twocolumn]{svjour3}          
\smartqed  

\usepackage[colorlinks,linkcolor=red,anchorcolor=blue,citecolor=blue,CJKbookmarks=True]{hyperref}
\usepackage{epsfig}
\usepackage{times}
\usepackage{url}            
\usepackage{booktabs}       
\usepackage{nicefrac}       
\usepackage{microtype}      
\usepackage{xspace}
\usepackage{graphicx}
\usepackage{wrapfig}
\usepackage{bm}
\usepackage{tabularx}
\usepackage{paralist}
\usepackage{natbib}
\usepackage{bbding}
\usepackage[algo2e]{algorithm2e}
\usepackage{afterpage}

\usepackage{multirow}%
\usepackage{amsmath,amssymb,amsfonts}%

\usepackage{amsthm}%
\usepackage{mathrsfs}%
\usepackage[title]{appendix}%
\usepackage{xcolor}%
\usepackage{textcomp}%
\usepackage{manyfoot}%
\usepackage{booktabs}%
\usepackage{algorithm}%
\usepackage{algorithmicx}%
\usepackage{algpseudocode}%
\usepackage{listings}%
\usepackage{pifont}
\usepackage{adjustbox}

\newcommand{\ie}{\emph{i.e.}}
\newcommand{\eg}{\emph{e.g.}}

\newcommand{\Tref}[1]{Table~\ref{#1}}

\newcommand{\Fref}[1]{Fig.~\ref{#1}}
\newcommand{\Sref}[1]{Sec.~\ref{#1}}

\begin{document}

\title{Affective Image Editing: Shaping Emotional Factors via Text Descriptions}

\author{Peixuan Zhang$^\dag$, Shuchen Weng$^\dag$, Chengxuan Zhu, Binghao Tang, Zijian Jia, \\ Si Li~\Envelope, Boxin Shi}

\institute{
$\dag$ Equal contribution.\vspace{0.1cm} \\
\Envelope \, Si Li (Corresponding author)\at
\email{lisi@bupt.edu.cn} \vspace{0.1cm}\\
\and
Peixuan Zhang \and Binghao Tang \and Zijian Jia \and Si Li \at
School of Artificial Intelligence, Beijing University of Posts and Telecommunications, China.
\and
Shuchen Weng\at
Beijing Academy of Artificial Intelligence
\and
Chengxuan Zhu \and Boxin Shi\at
State Key Laboratory for Multimedia Information Processing and National Engineering Research Center of Visual Technology, School of Computer Science, Peking University, China. \\
}

\date{Received: date / Accepted: date}

\maketitle
\begin{abstract}
In daily life, images as common affective stimuli have widespread applications.
Despite significant progress in text-driven image editing, there is limited work focusing on understanding users' emotional requests.
In this paper, we introduce \textbf{AIEdiT} for \textbf{A}ffective \textbf{I}mage \textbf{Edi}ting using \textbf{T}ext descriptions, which evokes specific emotions by adaptively shaping multiple emotional factors across the entire images.
To represent universal emotional priors, we build the continuous emotional spectrum and extract nuanced emotional requests.
To manipulate emotional factors, we design the emotional mapper to translate visually-abstract emotional requests to visually-concrete semantic representations.
To ensure that editing results evoke specific emotions, we introduce an MLLM to supervise the model training.
During inference, we strategically distort visual elements and subsequently shape corresponding emotional factors to edit images according to users' instructions.
Additionally, we introduce a large-scale dataset that includes the emotion-aligned text and image pair set for training and evaluation.
Extensive experiments demonstrate that AIEdiT achieves superior performance, effectively reflecting users' emotional requests.

\end{abstract}

\section{Introduction}
Affective images have significant potential to convey personal feelings and thoughts~\citep{yang2021stimuli}. 
As the common affective stimuli, affective images enable individuals from diverse backgrounds to understand emotional intentions between each other~\citep{decade}. 
Consequently, they are widely used in films, advertisements, and social networks, enhancing the appeal and attractiveness for content creators~\citep{chowdhury2008affective, uhrig2016emotion}. 
These widespread applications motivate researchers to explore approaches to further reduce the skill barriers and costs of creating affective images, empowering a broader audience to create and modify emotional visual content.

\begin{figure*}[!t]
\centering
    {
   \begin{minipage}[b]{\linewidth}
      \centering
     \includegraphics[width=1\linewidth]{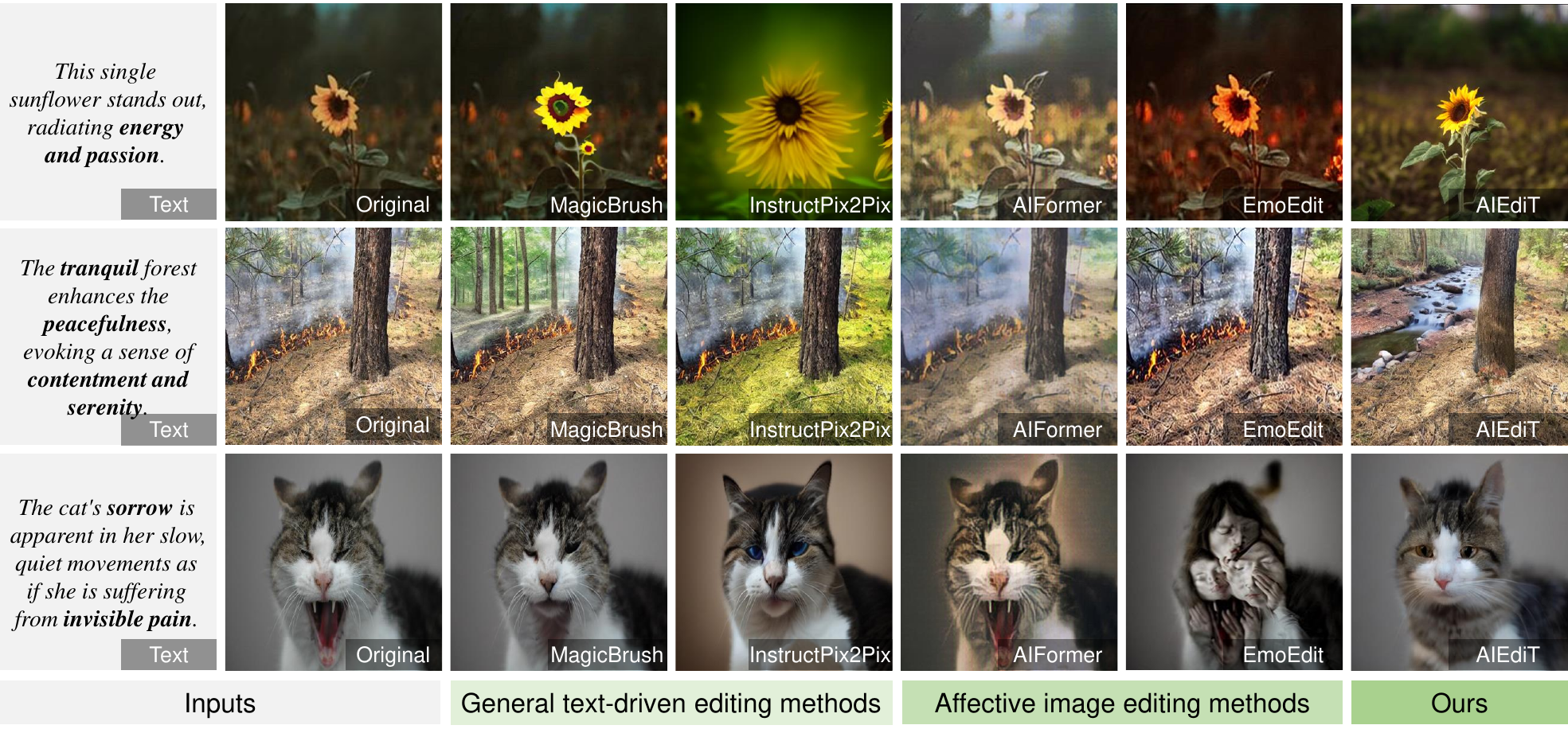}
  \end{minipage}%
    }
    \caption{Illustration of the proposed AIEdiT method: Given text descriptions that express users' emotional requests, AIEdiT creates affective images that evoke specific emotions through fine-grained text descriptions. In comparison, general text-driven editing methods (\eg, MagicBrush~\citep{magicbrush} and InstructPix2Pix~\citep{instructpix2pix}) fail to translate users’ emotional requests. Meanwhile, relevant affective image editing methods either focus on limited emotional factors (\eg, color tune in AIFormer~\citep{aif}) or use coarse-grained control instructions (\eg, emotion categories in EmoEdit~\citep{emoedit}).
    }
\label{fig:teaser}
\end{figure*}

Although existing general text-driven image editing methods~\citep{smartbrush, controlnet} have made notable improvements in controllability to synthesize visually-pleasing results, they still struggle to understand users' emotional requests~\citep{mlmgie} due to the inherent subjectivity and ambiguity of emotions~\citep{kim2003inverse,neta2021dynamic, neta2009corrugator}.
While recent affective image editing methods can effectively synthesize emotionally faithful results by integrating emotional priors, they primarily focus on editing limited emotional factors (\eg, color tune~\citep{aif}) or use coarse-grained control instructions (\eg, emotion categories~\citep{C2A2, emogen, happier, emoedit}). This restricts the options available for users to express nuanced individual emotions.
Therefore, there remains a significant challenge in designing a general affective image editing method that allows for comprehensive manipulation of emotional factors through a fine-grained interactive approach.

In this paper, we introduce \textbf{AIEdiT}, a method for \textbf{A}ffective \textbf{I}mage \textbf{Edi}ting using \textbf{T}ext descriptions. 
AIEdiT could create affective images that evoke specific emotions by adaptively shaping multiple emotional factors (\eg, color tune, object categories, and facial expression) across the entire image according to users' instructions.
By leveraging the interactivity and flexibility of text descriptions, users are able to freely describe their nuanced emotions, allowing AIEdiT to effectively translate emotional requests into images, as shown in \Fref{fig:teaser}: 
In the first row, AIEdiT adjusts the color tone by making the original images brighter;
In the second row, AIEdiT modifies object categories by transforming the fire into a tranquil stream, creating an atmosphere of contentment;
In the last row, AIEdiT controls the facial expression, changing the cat from angry to sorrowful.

We build AIEdiT based on the pre-trained cross-modality generative model~\citep{stablediffusion}, including a text encoder~\citep{clip} to capture complex semantics and diffusion models~\citep{ddpm,ddim} to create photo-realistic images. 
To represent universal emotional priors encapsulated in text descriptions, we additionally adopt BERT~\citep{bert} to extract emotional requests and build the emotional spectrum by learning a continuous emotional representation in a contrastive learning manner.
An emotional mapper is further designed to translate visually-abstract emotional requests to visually-concrete semantic representations, which consists of a stack of Transformer blocks equipped with self-attention and cross-attention mechanisms.
To ensure the created images evoke specific emotions, we take the Multimodal Large Language Model (MLLM)~\citep{sharegpt4v} as the supervisor, providing robust guidance for emotional understanding and consistency. 
During inference, we strategically add noise to distort visual semantic representations and subsequently shape the corresponding emotional factors according to user-provided text descriptions, finally producing emotionally faithful results.

To provide training data annotated with nuanced emotions, we extend existing EmoSet~\citep{emoset} into the \textbf{EmoTIPS} dataset, a high-quality \textbf{Emo}tion-aligned \textbf{T}ext and \textbf{I}mage \textbf{P}air \textbf{S}et, ensuring rich and diverse emotional annotation.
We adopt three validation criteria and conduct four human evaluation experiments to ensure the emotional consistency between texts and images. 
Four quantitative metrics are further designed to comprehensively evaluate the quality of image editing.

In summary, our contributions are listed as follows: 
\begin{itemize}
    \item We propose AIEdiT for affective image editing, which uses emotional text descriptions to shape emotional factors and create emotionally faithful results.
    \item We build the emotional spectrum to represent universal emotional priors and design the emotional mapper for translating emotional requests into visual elements.
    \item We take an MLLM as the supervisor of aligning created images with multiple emotional factors. Additionally, we introduce the EmoTIPS dataset for model training.  
\end{itemize}

\section{Related work}\label{sec:related}

\subsection{Multi-modal Sentiment Analysis}
Sentiment analysis is becoming increasingly significant in both natural language processing (NLP) and computer vision (CV) fields~\citep{zhang2018deep}. 
In NLP, emotion analysis of text is typically performed at the word level~\citep{ito2020word} and aspect level~\citep{zhang2022survey}, leveraging extracted textual features to accurately categorize sentiments. 
In CV, early approaches to image emotion analysis focus on separately analyzing color and textures~\citep{zhao2014exploring, zhao2014affective}, object categories~\citep{yao2020adaptive, rao2020learning}, and facial expression~\citep{highlevel1,highlevel2,highlevel4, yang2023context} properties of user-provided images. 
Based on these foundational studies, recent research~\citep{rao2016multi,wang2013interpretable} is able to integrate multiple emotional factors and therefore achieve a more comprehensive understanding of image emotions.
The advent of large language models (LLMs) demonstrates significant success in text understanding and response generation~\citep{llama2, gpt4v}. Researchers~\citep{anand2023multi} are encouraged to leverage the extensive prior knowledge embedded in LLMs to enhance text emotion understanding. 
To further improve the interaction between images and text, multimodal adapters~\citep{llava} and multi-layer perceptions~\citep{sharegpt4v} are applied to give LLMs multimodal processing capabilities. This advancement shows great potential for image emotion analysis and the joint sentiment analysis of text and images~\citep{li2024enhancing,li2023large,emollms}.

\subsection{Text-driven Image Editing}
Since text is a flexible and user-friendly medium that accurately expresses users' intentions, it has been widely used to guide image editing goals. The first attempts to use text as a condition are focused on generating text-related images for specific object categories~\citep{attnGAN, stackGAN}. With the introduction of large-scale image-text datasets~\citep{laion} and high-quality generative models~\citep{ddpm,ddim}, recent cross-modality generation models have the ability to generate a variety of scenarios~\citep{stablediffusion, imagen}. 
This inspires researchers to leverage pre-trained generative priors, and further explore approaches to edit user-provided images by fine-tuning parameters~\citep{multiconcept,dreambooth}, proposing additional adapters~\citep{smartbrush,controlnet}, and utilizing zero-shot, training-free strategies~\citep{sdedit, pix2pixzero}. As a result, text-driven image editing methods cover major generative research areas, \eg, low-level image processing~\citep{lcad, diffusingcolors} and high-level image translation~\citep{prompt2prompt, pix2pixzero}.
To enable models to better understand users' emotional requests, researchers collect large-scale emotional datasets with detailed text descriptions~\citep{artemis, artemisv2} and propose AIFormer~\citep{aif} to modify specific properties of affective images.
Although recent studies explore approaches using discrete emotion categories~\citep{emogen, happier, emoedit, C2A2} to manipulate more comprehensive emotional factors, the coarse-grained nature of emotion categories limits users' ability to express their nuanced emotional requests, where text descriptions continue to offer distinct advantages.

\section{EmoTIPS Dataset} \label{sec:dataset}
Although numerous affective image samples annotated with emotion categories are available~\citep{you2016building,liu2022ser30k}, they oversimplify emotional priors, making them inadequate to represent nuanced emotions. Additionally, existing datasets with emotional text descriptions~\citep{artemis,artemisv2} lack sufficient samples (over 100K), posing challenges for training a general model based on the limited emotional knowledge included.
As a result, we introduce the \textbf{EmoTIPS} dataset\footnote{The dataset will be released once the paper is published.}, an \textbf{Emo}tion-aligned \textbf{T}ext and \textbf{I}mage \textbf{P}air \textbf{S}et, to enable image editing models to effectively understand users' emotional requests and achieve comprehensive manipulation of emotional factors. We collect affective images from the EmoSet~\citep{emoset} and generate corresponding emotional text descriptions for each sample using the MLLM~\citep{sharegpt4v}. 
Specifically, the pipeline for dataset construction includes three phases: annotation, validation, and evaluation, as detailed below.

\begin{figure}[t]
   \centering
  \includegraphics[width=\linewidth]{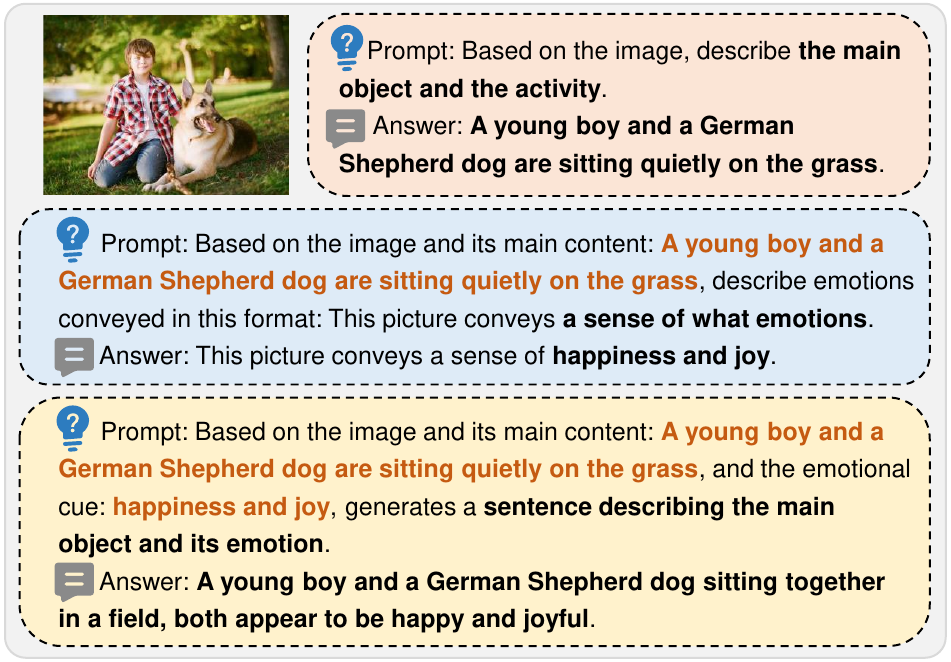}
  \caption{Visualization of the annotation process which a Chain-of-Thought prompting strategy.
  }
  \label{fig:annotation_dataset}
\end{figure}

\noindent \textbf{Annotation.} 
Due to the inherent subjectivity and ambiguity of emotions, we employ a Chain-of-Thought prompting strategy~\citep{kojima2022large} to guide the MLLM in accurately understanding and describing emotions step by step: 
\textit{(i)} prompting the MLLM to identify the activities of the main objects in the image; 
\textit{(ii)} incorporating the detected content as additional prompts to have the MLLM estimate emotional cues within the image; and
\textit{(iii)} combining the previously estimated image content and emotional cues to prompt the generation of full text descriptions.
Notably, while constructing the dataset, we intentionally exclude descriptions of object appearances that can directly express emotions (\eg, colors, styles, and facial expressions), thereby focusing the text descriptions exclusively on emotional cues.
As a result, the annotated text descriptions include only the main objects and their corresponding emotions. We present an annotation sample and prompts used in the Chain-of-Thought prompting strategy in \Fref{fig:annotation_dataset}.

\begin{figure}[t]
   \centering
  \includegraphics[width=\linewidth]{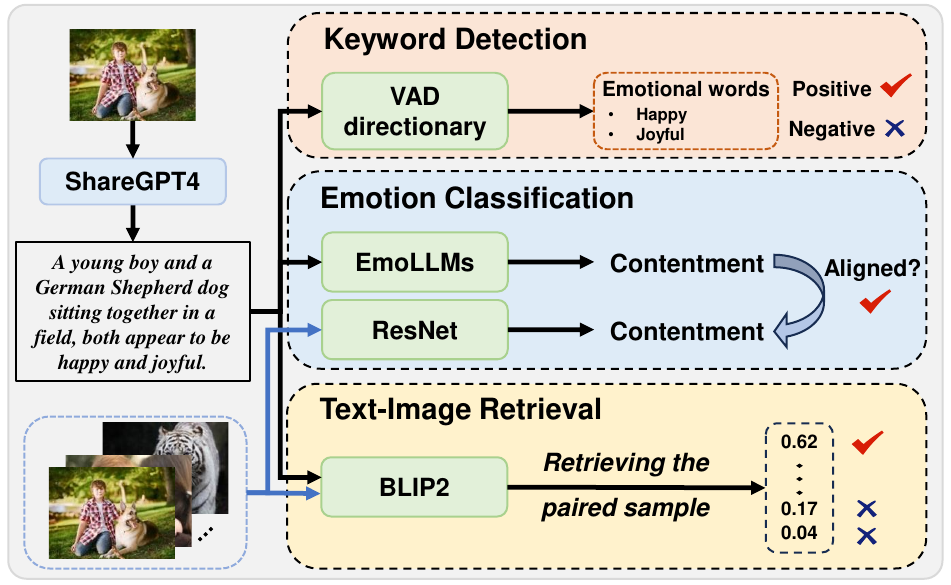}
  \caption{Visualization of the three validation criteria: keyword detection, emotion classification, and text-image retrieval.}
  \label{fig:validation_dataset}
\end{figure}

\noindent \textbf{Validation.}
Due to the nature of the LLMs that generate diverse responses, we adopt three validation criteria to ensure the generated texts accurately describe the emotional cues within the images:
\textit{(i)} Keyword Detection. We utilize the VAD dictionary~\citep{vad} to classify emotional words into binary categories of positive and negative. Samples that do not include words matching the sentiment polarity of the images are discarded.
\textit{(ii)} Emotion Classification. We utilize the emotional LLMs~\citep{emollms} and image classifier~\citep{resnet} to estimate the emotion categories on the Mikel's wheel~\citep{mikelwheel}. 
Samples where the categories do not align are discarded.
\textit{(iii)} Text-Image Retrieval. We randomly select $127$ additional images from the dataset and use a pre-trained text-image retrieval model~\citep{li2023blip}. 
Samples are discarded if the paired image does not rank within the top 10 retrieval results.
For clarify, we visualize the annotation pipeline and aforementioned three validation criteria in \Fref{fig:validation_dataset}.

\noindent \textbf{Evaluation.}
We conduct four human evaluation experiments to assess the quality of our annotated dataset, focusing on whether \textit{(i)} the images are emotionally expressive; \textit{(ii)} the text descriptions are emotionally expressive; \textit{(iii)} the images and text descriptions share similar emotional expression; and \textit{(iv)} the text descriptions accurately describe the content of the images. For each experiment, 25 volunteers are asked to evaluate 100 random samples given the choices of ``Failed'', ``Borderline'', ``Acceptable'' and ``Perfect''.
As shown in \Tref{tab:data_evaluation}, in each experiment, over 90\% volunteers rate the annotation quality as ``Acceptable'' or higher, demonstrating the robustness of our dataset.

\begin{table}[t]
\centering
\setlength\tabcolsep{11pt}

\caption{Percentage (\%) of user ratings in the four experiments of human evaluation for the EmoTIPS dataset. Throughout the paper, best scores are highlighted in \textbf{bold}.}
\begin{adjustbox}{width={0.48\textwidth},totalheight={\textheight},keepaspectratio}
\begin{tabular}{lcccc} \cr \toprule
Rating & Exp-I & Exp-II & Exp-III & Exp-IV  \\ \midrule
    Failed & $0.00$ & $0.00$ & $1.72$ & $0.56$   \\ 
    Borderline & $1.76$ & $3.04$ & $7.24$ & $5.12$   \\ 
    Acceptable & $9.64$ & $21.68$ & $32.56$ & $26.60$   \\ 
    Perfect & $\mathbf{88.60}$ & $\mathbf{75.28}$ & $\mathbf{58.48}$ & $\mathbf{67.72}$   \\ 
    \bottomrule
\end{tabular}\label{tab:data_evaluation}
\end{adjustbox}
\vspace{2mm}
\end{table}

\begin{figure*}[h]
  \centering
  \includegraphics[width=\linewidth]{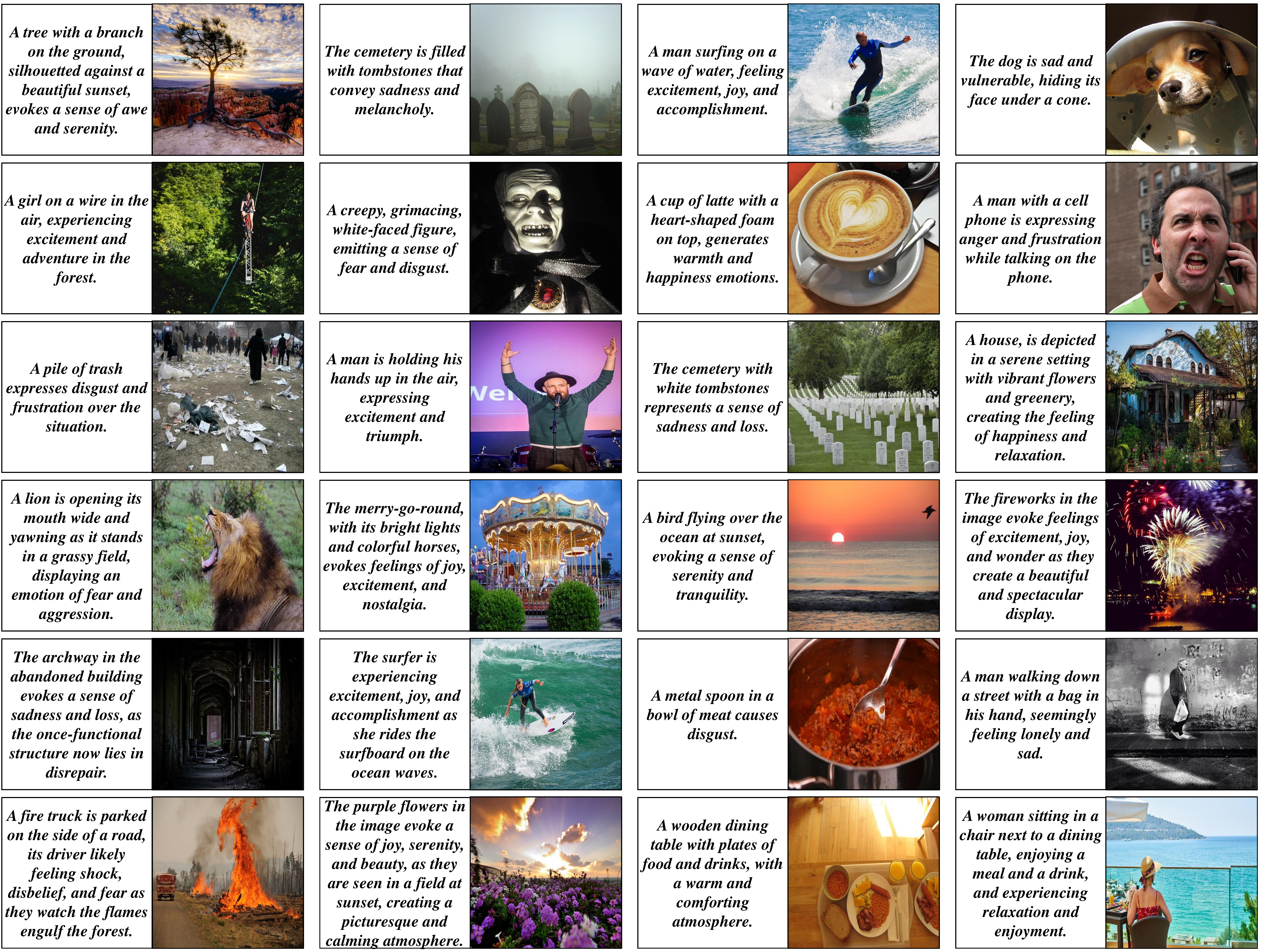}
  \caption{Randomly selected samples from the EmoTIPS dataset.}
  \label{fig:supp_dataset}
\end{figure*}

\noindent \textbf{Summary.}
The introduced EmoTIPS dataset includes 1M affective images sourced from EmoSet~\citep{emoset}, with each image paired with a corresponding emotional text description.
Additionally, we have reserved 3K samples specifically for model evaluation. 
Each sample consists of the original image, a target image that has a similar embedding~\citep{vgg}, an emotional text description of the target image, and an emotional category distribution of the target image.
During evaluation, the model is only provided with the original image and an emotional text description as the editing instruction. The emotional category distribution of the target image serves as the ground truth for calculating quantitative scores. 
Additionally, we show randomly selected samples from the dataset in~\Fref{fig:supp_dataset} to present the diversity. 

\begin{figure*}[t]
   \centering
  \includegraphics[width=\linewidth]{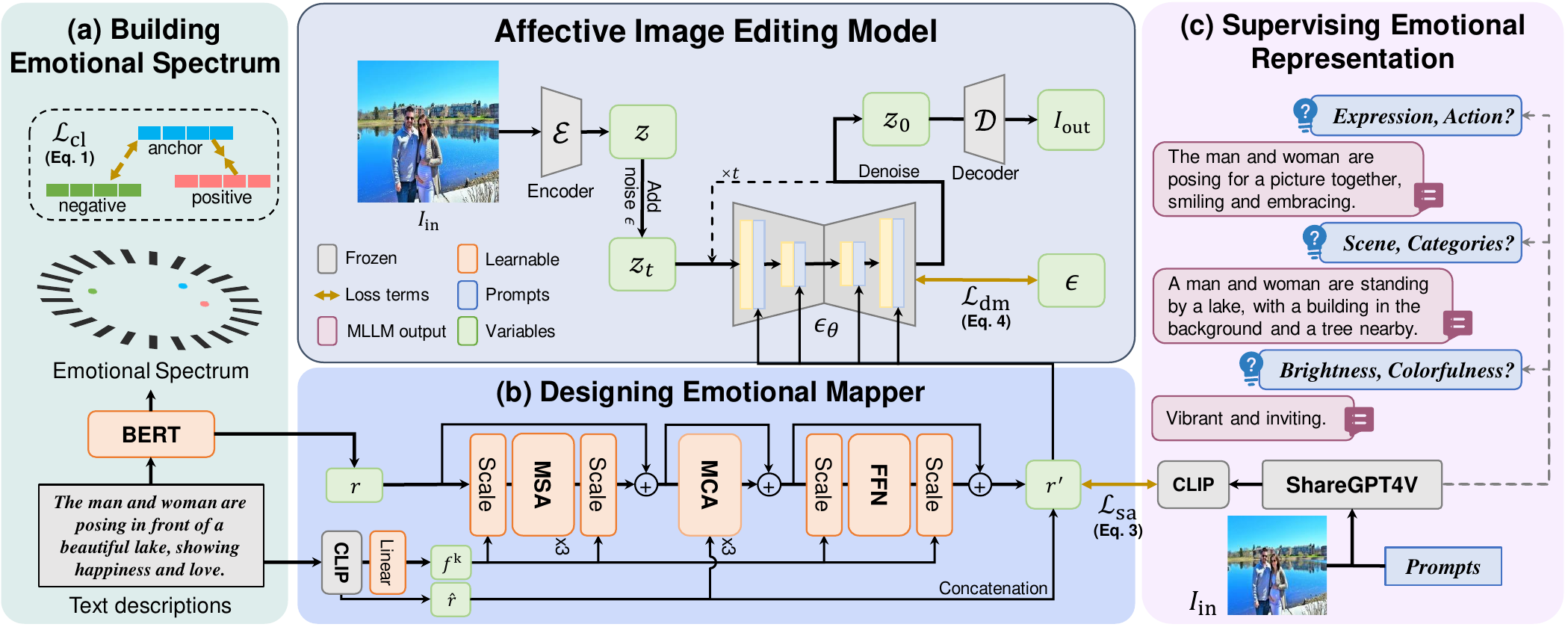}
  \caption{Overview of the proposed AIEdiT model. 
  (a) To represent universal emotional priors encapsulated in text descriptions, we build the continuous emotional spectrum and extract nuanced emotional requests (\Sref{sec:emopair}); 
  (b) To manipulate emotional factors, We design the emotional mapper to translate visually-abstract emotional requests into visually-concrete semantic representations (\Sref{sec:understanding});
  (c) To ensure specific emotions are effectively evoked, we introduce the MLLM to supervise emotional understanding (\Sref{sec:mllm});
  During inference, we strategically distort the visual semantic representations and shape the corresponding emotional factors unser the guidance of text descriptions (\Sref{sec:editing}).
  The emotional mapper is designed as a multi-modal Transformer, facilitating global interaction among input conditions, composed of a stack of Multi-head Self-Attention (MSA) blocks, Multi-head Cross-Attention (MCA) blocks, and Feed-Forward Networks (FFN).
  During the training, we keep the latent diffusion model frozen to preserve its generative ability.
  }
  \label{fig:pipeline}
\end{figure*}

\section{Methodology}
In this section, we first build the continuous emotional spectrum to represent universal emotional priors encapsulated in text descriptions (\Sref{sec:emopair}). Then, we design an emotional mapper that translates visually-abstract emotions into visually-concrete semantic representations (\Sref{sec:understanding}). Next, we introduce a multi-modal large language model to supervise emotional understanding, which ensures the created images accurately evoke specific emotions (\Sref{sec:mllm}).
Based on this mapping relationship between abstract emotional requests and concrete emotional factors, we present a strategy for editing affective images by manipulating emotional factors across the entire image according to users' instructions~(\Sref{sec:editing}), along with the training details (\Sref{sec:detail}).

\subsection{Building Emotional Spectrum} \label{sec:emopair}
Previous works~\citep{happier, emoedit} manipulate the emotional factors of affective images using discrete emotion categories. However, such coarse-grained control instructions limit users' ability to express nuanced emotional requests, resulting in reduced controllability when shaping emotional factors (\eg, the emotion of awe can be conveyed through the brilliance of fireworks or the grandeur of mountains). This motivates us to represent emotional priors encapsulated in text descriptions and build a continuous emotional spectrum, enabling flexible and accurate correspondences between users' requests and emotional factors.

As presented in \Fref{fig:pipeline}~(a), we first adopt BERT~\citep{bert} as the text encoder to extract emotional requests $r \in \mathbb{R}^{C^{\mathrm{t}} \times N^{\mathrm{l}}}$, where $C^{\mathrm{t}}$ is the number of embedding channels and $N^{\mathrm{l}}$ is the length of text descriptions. 
Since affective images evoking similar emotions often share common emotional expressions, we first estimate the emotional distribution for each image. Specifically, we use a ResNet~\citep{resnet} pre-trained on EmoSet~\citep{emoset} to estimate the distribution $d \in \mathbb{R}^{N^{\mathrm{c}}}$ over the $N^{\mathrm{c}}$ emotion categories defined by Mikel's wheel~\citep{mikelwheel}.
Next, we create samples $s=(r,d)$ by pairing each emotional request $r$ with the corresponding emotional distribution $d$. 
We further define relationships between these samples based on positions defined by the Mikel's wheel~\citep{mikelwheel}, where emotions located in the same region are considered positive pairs, while emotions located opposite each other (\eg, awe and disgust) are treated as negative pairs.
Therefore, we sample $N^{\mathrm{p}}$ such pairs using the batch-hard mining \citep{triplet_loss}, structured as tuples $[s^{\mathrm{anc}}_{i}, s^{\mathrm{pos}}_{i}, s^{\mathrm{neg}}_{i}]$, where $i \in \{1,\dots,N^{\mathrm{p}}\}$ is the index, $s^{\mathrm{anc}}$, $s^{\mathrm{pos}}$, and $s^{\mathrm{neg}}$ represent anchor, positive, and negative pairs, respectively.
After that, we build the emotional spectrum by learning a continuous emotional representation: 
\begin{equation}
\!\! \mathcal{L}_{\mathrm{cl}} \!=\! \sum_{i=1}^{N^{\mathrm{p}}} \mathrm{max} \big( 0, \mathrm{dis}(s^{\mathrm{anc}}_i, s^{\mathrm{pos}}_i) - \mathrm{dis}(s^{\mathrm{anc}}_i, s^{\mathrm{neg}}_i) + \alpha \big), \!\!
\end{equation}
where $\mathrm{dis}(s_{i}, s_{j})  = \|r_{i} - r_j\|_2/ \|d_{i} - d_j\|_2$ is a distance function that measures the sentiment similarity between pairs and $\alpha=0.2$ is a hyper-parameter that controls the emotional margin.
When forming training tuples, we randomly select samples from the dataset.
The loss reduces the emotional distance between samples with similar emotions while increasing the distance between those that evoke opposite emotions, allowing the emotional spectrum to reflect emotions across all emotional dimensions.
After building the emotional spectrum, the BERT encoder can extract emotional requests encapsulated in text descriptions and remain frozen.

\subsection{Designing Emotional Mapper} \label{sec:understanding}
With the emotional spectrum constructed, emotional requests can be extracted from text descriptions. However, there remains a challenge in accurately translating these visually-abstract emotional requests into visually-concrete semantic representations. Therefore, we design the emotional mapper to address this issue.

As illustrated in \Fref{fig:pipeline}~(b), we employ a pre-trained CLIP model~\citep{clip} to extract text semantics $\hat{r} \in \mathbb{R}^{C^{\mathrm{s}} \times N^{\mathrm{l}}}$, where $C^{\mathrm{s}}$ is the number of embedding channels. 
Given that CLIP aligns visual and textual representations, we intend to further translate text semantics into emotion-aligned visual elements.
To achieve this, we introduce a linear layer to adaptively extract the key semantic features $f^{\mathrm{k}} \in \mathbb{R}^{C^{\mathrm{s}}}$. These features, along with the emotional requests $r$ and text semantics $\hat{r}$, are fed into a multi-modal Transformer. This Transformer consists of a stack of Multi-head Self-Attention blocks (MSA) to capture overall emotional semantics, Multi-head Cross-Attention blocks (MCA) to integrate emotional semantics with related visual elements, and Feed-Forward Networks (FFN) to deeply understand the translation of text semantics. Throughout the translation process, the key semantic features $f^\mathrm{k}$ are used to scale the emotional requests after each sub-module as:
\begin{equation}
    \hat{f}^{\mathrm{r}} = \big (1 + \mathbf{W}_1 f^{\mathrm{k}} \big) \odot \frac{f^{\mathrm{r}}-\mu}{\sigma} + \mathbf{W}_2 f^{\mathrm{k}}, 
    \label{eq:spade}
\end{equation}
where $f^\mathrm{r}$ and $\hat{f}^{\mathrm{r}}$ represent the intermediate feature maps of emotional requests before and after scaling, respectively. $\mu$ and $\sigma$ denote the mean and standard deviation of $f^{\mathrm{r}}$, and $\odot$ means element-wise multiplication, while $\mathbf{W}_{\{1,2\}}$ are learnable matrices.
Finally, the emotional mapper outputs the visually-concrete semantic representation $r^\prime \in \mathbb{R}^{{C^{\mathrm{s}} \times N^\mathrm{l}}}$.

\begin{figure}[t]
  \centering
  \includegraphics[width=\linewidth]{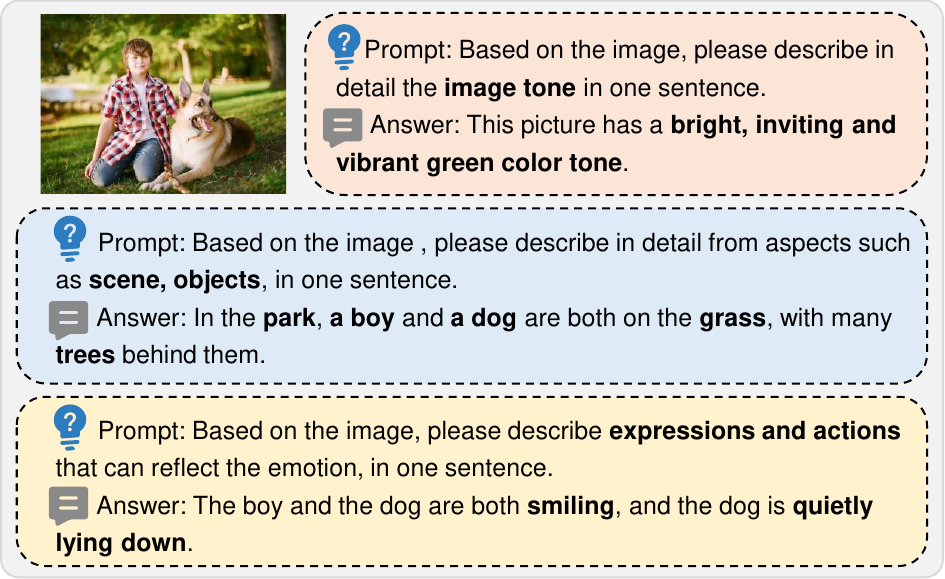}
  \caption{Visualization of MLLM prompts used for emotional supervision. }
  \label{fig:supp_prompt}
\end{figure}

\subsection{Supervising Emotional Understanding} \label{sec:mllm}
Emotional factors (\eg, color tune, object categories, and facial expression)~\citep{rao2019multi,rao2020learning} play a crucial role in how affective images evoke emotions. 
In the absence of target emotional images, the MLLM~\cite{sharegpt4v} is introduced to supervise the emotional mapper to effectively create visual elements with corresponding emotional factors.

As shown in \Fref{fig:pipeline}~(c), the MLLM is tasked with responding to our questions based on the image content and its comprehensive emotional understanding. We specifically design prompts for the MLLM to effectively analyze representative emotional factors and provide comprehensive responses.
After the MLLM responds to each pre-defined prompt, we extract the CLIP embeddings of the responses and apply a sentiment alignment loss to minimize the mean square error with the semantic representation generated by the emotional mapper:
\begin{equation}
\mathcal{L}_\mathrm{sa} = \sum_{i=1}^{N^\mathrm{r}} \|\phi(x^{\mathrm{t}}) - \varphi(x^{\mathrm{r}}) \|_2,
\end{equation}
where $\phi$ and $\varphi$ refer to the emotional mapper and text encoder in CLIP~\citep{clip}, respectively. 
$x^{\mathrm{t}}$ and $x^{\mathrm{r}}$ represent the user-provided text descriptions and responses from the MLLM separately, and $N^{\mathrm{r}}$ indicates the number of responses. 
Additionally, to maintain the nature of the diffusion model, we preserve the noise prediction loss:
\begin{equation}
\mathcal{L}_{\mathrm{dm}} = \mathbb{E}_{t, z, \epsilon \sim \mathcal{N}(0,1)} \big[\| \epsilon - \epsilon_{\theta} \big(z_{t}, t, \phi(x^{\mathrm{t}}) \big) \|_{2}\big], \label{eq:dm}
\end{equation}
where $\epsilon_{\theta}$ is the diffusion model and $t$ is the timestep in the denoising process. The total training loss is a combination:
\begin{equation}
\mathcal{L}_\mathrm{total} = \mathcal{L}_\mathrm{sa} + \beta \mathcal{L}_{\mathrm{dm}}, \label{eq:totalloss}
\end{equation}
where $\beta=10$. 
In practice, we only fine-tune the designed emotional mapper, and keep the backbone latent diffusion model and the autoencoder frozen, to preserve the generative ability of pre-trained models. 

For better reproducibility of our results, We present the prompts used for supervision in \Fref{fig:supp_prompt}. Notably, directly replacing the emotional mapper with the MLLM is infeasible. This is because MLLMs require visual input to extract relevant cues, whereas during editing, the target image is unavailable, and cues have to be derived solely from the user-provided text descriptions. Instead, we train the emotional mapper via self-reconstruction, thereby avoiding the need for target images.

\begin{figure}[t]
  \centering
  \includegraphics[width=\linewidth]{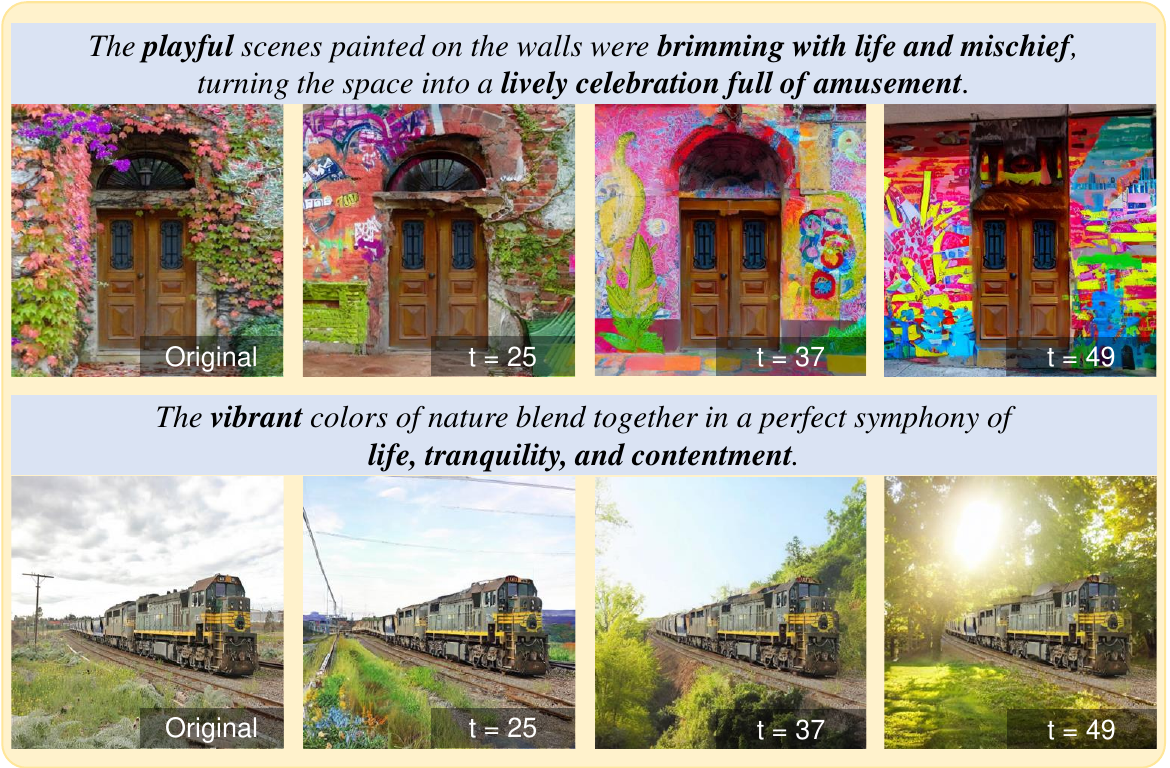}
  \caption{Visualization of the correspondences between manipulated emotional factors and the number of noise-adding steps $t=25,37,49$.}
  \label{fig:application_t}
\end{figure}

\subsection{Editing Affective Images} \label{sec:editing}
With the emotional mapper, the model effectively interprets emotional requests and translates them into visual elements aligned with the intended emotions. To further leverage the emotional knowledge learned by the emotional mapper and the generative prior preserved in diffusion models, we propose the editing strategy to modify affective images during the inference process.

Technically, for the user-provided affective image $I_{\mathrm{in}}$, we adopt a pre-trained image encoder $\mathcal{E}$ to map it into the latent space as $z = \mathcal{E}(I_{\mathrm{in}})$. 
Then, we strategically distort the visual semantic representations of the latent code by adding appropriate noise.
Since the forward process of diffusion models~\citep{ddpm, ddim} can be expressed as a linear combination of the original latent code and a noise variable $\epsilon  \sim \mathcal{N}(0, 1)$, we can directly calculate the noised latent code distorted by adding noise $t$ times as $z_{t} = \sqrt{\alpha_{t}} z + \sqrt{1-\alpha_{t}} \epsilon$, where $\alpha_{t}$ is a hyperparameter. 
Next, we iteratively refine the distorted latent code and shape corresponding emotional factors under the guidance of text descriptions as
$z_{t-1} = 1/\sqrt{a_{t}} \cdot \big (z_{t} - \epsilon_\theta (z_{t}, t, \phi({x^{\mathrm{t}}})) \cdot (1-\alpha_{t})/{\sqrt{1-\bar{\alpha}_{t}}}  \big ) + \sigma_{t}\epsilon$, where $\epsilon_\theta$ is our denoising network equipped with the emotional mapper, $\bar{\alpha}_{t}=\smash{\prod_{s=1}^{t}\alpha_{s}}$ is the cumulative noise amount, and $\sigma_{t}$ is a hyperparameter. After $t$ iterations, we obtain the restored latent code $z_{0}$.
Finally, a pre-trained image decoder $\mathcal{D}$ maps the edited latent to the image domain as $I_{\mathrm{out}} = \mathcal{D}(z_{0})$.

As illustrated in \Fref{fig:application_t}, we observe that as the amount of added noise increases, the edited visual features change from color tone to object category, and eventually to overall semantics. This progression corresponds to emotional factors ranging from low to high levels~\citep{decade}, indicating that the emotional mapper translates emotional requests following the hierarchical structure of emotional factors.

\subsection{Training Details} \label{sec:detail}
We initialize the backbone with SD1.5 parameters, keeping them frozen to maintain generative priors. Training is performed on two NVIDIA GeForce 3090 GPUs using the Adam optimizer \citep{kingma2014adam} at a learning rate of $5 \times 10^{-5}$. Initially, we spend 36 hours constructing the emotional spectrum, followed by an additional 96 hours optimizing the emotional mapper with emotional spectrum frozen.

\section{Experiment}

\subsection{Quantitative Evaluation Metrics}
We utilize three quantitative metrics to evaluate the performance of relevant methods. For each metric, we detail what it measures and how it is calculated below:

\begin{itemize}
\item The \textbf{Fréchet Inception Distance (FID)}~\citep{FID} is a metric used to evaluate whether edited images share a similar distribution with real images. Specifically, we utilize the pre-trained Inception-v3 network~\citep{Inception} to extract features from both the editing and real images, and calculate the mean and covariance of the extracted features to create corresponding multivariate Gaussians. Finally, we measure the Fr\'echet distance between these Gaussians to assess the distribution similarity between editing and real images 
to reflect visual quality.

\item The \textbf{Sematic Clarity (Sem-C)} is a metric used to evaluate whether the edited images present clearly recognizable visual contents to evoke corresponding emotions.
Following EmoGen \citep{emogen}, we categorize the edited images using the pre-trained object classifier from ImageNet~\citep{imagenet} and the scene classifier from PLACES365~\citep{place365}.
This score is then calculated as the average of the highest probability assigned by either classifier across all images.

\item The \textbf{Kullback-Leibler Divergence (KLD)}~\citep{kullback1951information} is a metric used to evaluate the matching accuracy between editing results and emotional requests. Specifically, we use the fine-tuned ResNet-50~\citep{resnet} to categorize the editing results and target images into emotional categories based on Mikel's wheel~\citep{mikelwheel}, forming corresponding emotional distributions. After that, we calculate the Kullback-Leibler divergence between them to assess the emotional distance. Instead of directly calculating the accuracy, KLD provides a more nuanced evaluation of whether the target emotions are effectively evoked.
\end{itemize}

Additionally, we present \textbf{user preference (Pref.)} for the edited images of each method through a user study, demonstrating whether the created images align with the user-provided text descriptions.

\afterpage{
    \begin{figure*}[p]
        \centering
        \includegraphics[width=\linewidth]{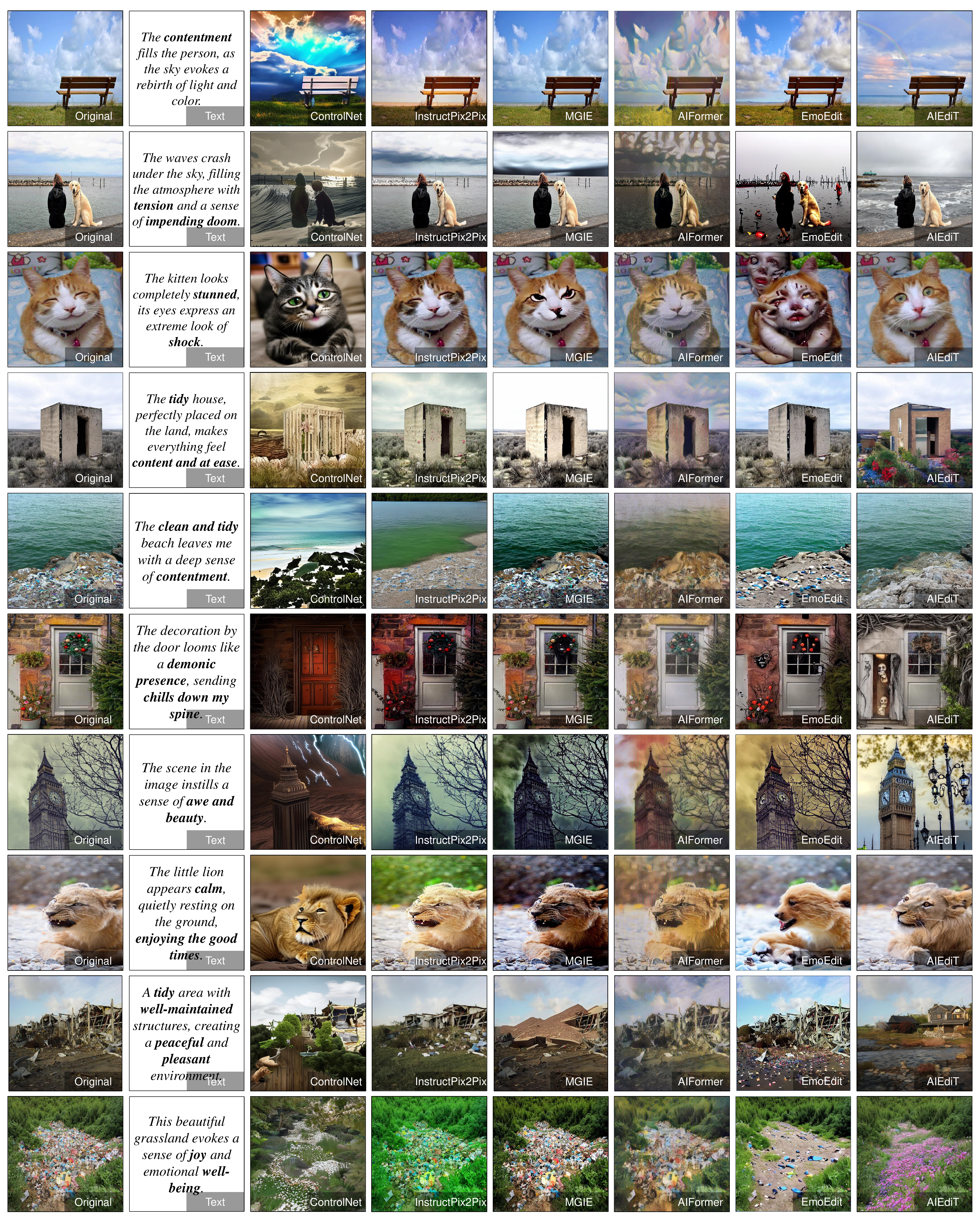}
        \caption{Visual quality comparisons with general text-driven image editing methods and affective image editing methods. AIEdiT could understand emotional requests and shape corresponding emotional factors based on text descriptions.}
        \label{fig:comparison}
    \end{figure*}
}

\subsection{Comparison with State-of-the-art Methods} \label{sec:comparison}

To demonstrate the advantages of our AIEdiT in shaping multiple emotional factors based on text descriptions, we compare it with state-of-the-art general text-driven image editing methods (\ie, ControlNet~\citep{controlnet}, InstructPix2Pix \citep{instructpix2pix}, and MGIE~\citep{mlmgie}) and affective image editing methods (\ie, AIFormer~\citep{aif} and EmoEdit~\citep{emoedit}).

\noindent \textbf{Qualitative comparisons.}
In \Fref{fig:comparison}, we present visual quality comparisons with the aforementioned methods. 
Intuitively, general text-driven image editing methods struggle to understand the emotional requests. Specifically, ControlNet~\citep{controlnet} overly modifies the object appearance (\eg, \Fref{fig:comparison} first row, the unnatural blue sky glow);  
InstructPix2Pix~\citep{instructpix2pix} often presents fewer modifications (\eg, \Fref{fig:comparison} second row, the calm lake surface);
MGIE~\citep{mlmgie} struggle to enhance the visual appeal, (\eg, \Fref{fig:comparison} third row, the skewed eyes of cat);
On the other hand, affective image editing methods are limited by their controllability.
AIFormer~\citep{aif} can solely evoke specific emotions by modifying low-level features, it fails when the image content is ill-suited (\eg, \Fref{fig:comparison} fourth row, attempting to convey contentment within a desolate environment;
Although EmoEdit~\citep{emoedit} edits images based on emotional categories, unable to evoke fine-grained emotions encapsulated in text descriptions (\eg, \Fref{fig:comparison} fifth row, the beach garbage partially altered, but still disorganized).
In contrast, our AIEdiT accurately shapes corresponding emotional factors to reflect users' emotional requests.

\begin{figure*}[t]
  \centering
  \includegraphics[width=0.9\linewidth]{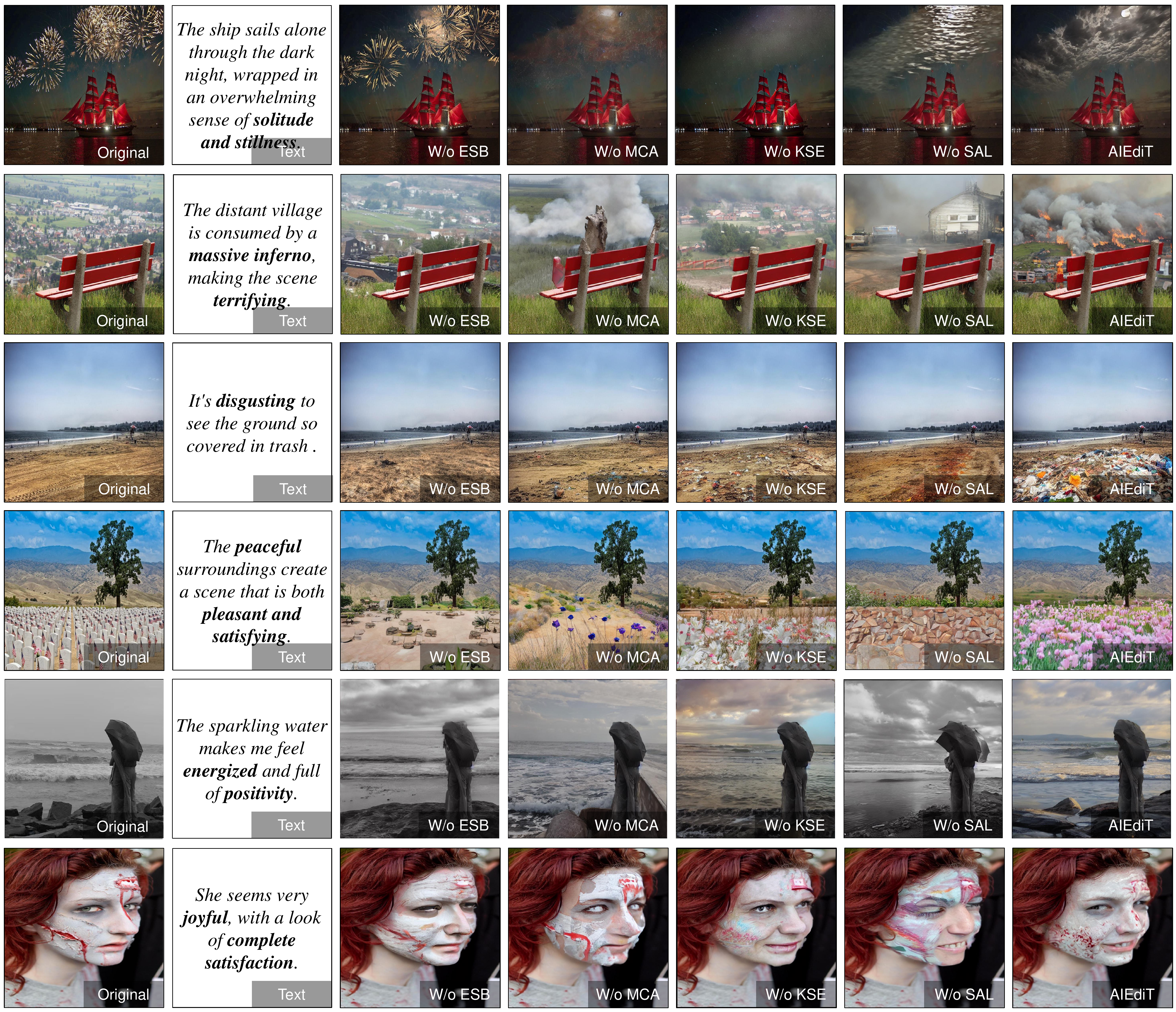}
  \vspace{2mm}
  \caption{Ablation study results with different variants of the AIEdiT. After removing the proposed modules, the created images cannot accurately reflect users' emotional requests.}
  \label{fig:ablation}
\end{figure*}

\noindent \textbf{Quantitative comparisons.}
We present quantitative comparison results in \Tref{tab:comparison} compared with general text-driven image editing methods~\citep{controlnet, instructpix2pix, mlmgie} and affective image editing methods~\citep{aif, emoedit}. This demonstrates that our method outperforms across all three quantitative metrics, demonstrating AIEdiT’s ability to create photo-realistic images (FID) with clearly recognizable visual content (Sem-C) that accurately reflect users' emotional requests (KLD).

\noindent \textbf{User study.}
To further assess the subjective preferences, We conduct a user study to evaluate whether our method is preferred by human observers.
In the experiment, participants are shown an original image, an emotional text description, and the corresponding results generated by the related general text-driven image editing methods and affective image editing methods. 
They are asked to select the result that best aligns with the emotional text descriptions. 
We conduct the user study on Amazon Mechanical Turk (AMT) using 100 samples randomly selected from the EmoTIPS dataset, with results polled from 25 volunteers.
As shown in \Tref{tab:comparison}, our method achieves the highest user preference, demonstrating 
its superior performance to produce affective images that closely align with user-provided emotional requests.

\subsection{Ablation Study} \label{sec:ablation}
We discard various modules and create four variants to study the impact of our proposed modules and designed losses. The evaluation scores and visual results of the ablation study are shown in~\Tref{tab:comparison} and ~\Fref{fig:ablation}, respectively.

\noindent \textbf{W/o ESB (Emotional Spectrum Building).} 
We discard the emotional spectrum built for the BERT. Without this emotional prior, this variant struggles to understand users' emotional requests (\eg, \Fref{fig:ablation} first row, fireworks are less relevant to a sense of solitude).

\noindent \textbf{W/o MCA (Multi-head Cross-Attention).} We remove all multi-head cross-attention blocks in the emotional mapper. 
Lacking the integration with related visual elements, the image content changes inaccurately (\eg, \Fref{fig:ablation} second row, wood inexplicably appears in front of the bench).

\noindent \textbf{W/o KSE (Key Semantic Extraction).} We disable the linear layer for key semantic extraction and provide scaling instructions. As a result, this ablation has difficulty fully translating emotional requests (\eg, \Fref{fig:ablation} third row, the beach becomes dirty but not sufficiently disgusting).

\noindent \textbf{W/o SAL (Sentiment Alignment Loss).} We discard the sentiment alignment loss, preventing the MLLM from supervision. 
Consequently, the model struggles to shape corresponding emotional factors (\eg, \Fref{fig:ablation} fourth row, transforming the cemetery into a brick wall is meaningless).

\begin{table}[t]
\centering
\caption{Quantitative experiment results of comparison and ablation. $\uparrow$ ($\downarrow$) means higher (lower) is better. } 
\setlength\tabcolsep{8pt}
\centering
\begin{adjustbox}{width={0.48\textwidth},totalheight={\textheight},keepaspectratio}
\begin{tabular}{l c c c c} \cr \toprule
\multirow{1}{*}{Method} 
& FID $\downarrow$ & Sem-C $\uparrow$   & KLD $\downarrow$  & Pref. (\%) $\uparrow$  \\ \midrule
    \multicolumn{5}{c}{Comparison with state-of-the-art methods} \cr \midrule
    ControlNet  & $32.58$  & $0.656$ & $2.7694$ & $9.12$  \\ 
    InstructPix2Pix & $32.77$ & $0.649$ & $2.8952$ & $6.36$  \\
    MGIE   & $29.82 $ & $0.655$ & $2.8302$ & $15.08$  \\ 
    AIFormer & $38.56$ & $0.526$ & $2.6940$ & $7.32$  \\ 
    EmoEdit & $37.10$ & $0.620$ & $2.6219$ & $22.44$  \\
    Ours (AIEdiT) & $\mathbf{27.93 }$ & $\mathbf{0.685}$ & $\mathbf{2.4373}$ & $\mathbf{39.68}$   \\ \midrule
    \multicolumn{5}{c}{Ablation study} \cr \midrule
    \textit{W/o} ESB & $29.67$ & $0.672$ & $2.6160$ & $N/A$  \cr
    \textit{W/o} MCA  & $28.98$ & $0.666$ & $2.4894$ & $N/A$ \cr
    \textit{W/o} KSE & $28.14$ & $0.657$ & $2.5139$ & $N/A$ \cr
    \textit{W/o} SAL & $31.03$ & $0.651$ & $2.5597$ & $N/A$ \cr\bottomrule

\end{tabular}\label{tab:comparison}
\end{adjustbox}
\vspace{1mm}
\end{table}

\begin{table}[t]
\centering
\caption{Quantitative experiment results for different hyper-parameters $t=25,37,49$.}\label{tab:editing_strategy}
\vspace{2mm}
\setlength\tabcolsep{10pt}
\begin{adjustbox}{width={0.48\textwidth},totalheight={\textheight},keepaspectratio}
\begin{tabular}{l|c c c c}
    \toprule
    Parameter & FID $\downarrow$ & Sem-C $\uparrow$   & KLD $\downarrow$  & Pref (\%) $\uparrow$  \\ \midrule
    $t$ = 25  & $\mathbf{20.40} $  & $0.625$ & $2.4805$ & $27.96$  \\ 
    $t$ = 37  & $27.93 $ & $ \mathbf{0.685}$ & $2.4373$ & $\mathbf{45.32}$  \\ 
    $t$ = 49  & $32.37$ & $ 0.666$ & $\mathbf{2.4301}$ & $26.72$  \\  
    \bottomrule
    \end{tabular}
\end{adjustbox}
\end{table}

\subsection{Hyper-parameter Analysis}
We evaluate the impact of the hyper-parameter $t$, which determines the amount of added noise in editing strategies (\Sref{sec:editing}). 
As previously illustrated qualitatively in \Fref{fig:application_t}, increasing $t$ leads the image manipulation to gradually evolve from low-level color tone towards high-level overall semantics to achieve the desired emotional expression. To further validate this relationship quantitatively, we present additional experiment results in \Tref{tab:editing_strategy}, which confirms this trend. 

Specifically, as the noise-adding step $t$ increases, edited images become greater levels of distortion and reduced visual fidelity compared to the originals. This decline in visual quality is reflected in worse FID scores, indicating a larger feature gap. Concurrently, higher noise levels prove more effective in shaping the corresponding emotional factors, leading to improved KLD scores that indicate a more accurate match to the user's emotional requests.
We observe that adding an optimal level of noise can produce clear visual content while maximizing emotional expression, resulting in the highest Sem-C and user study scores. Therefore, we select $t=37$ for all of our experiments to balance the visual quality and emotional fidelity.

\begin{figure}[t]
  \centering
  \includegraphics[width=\linewidth]{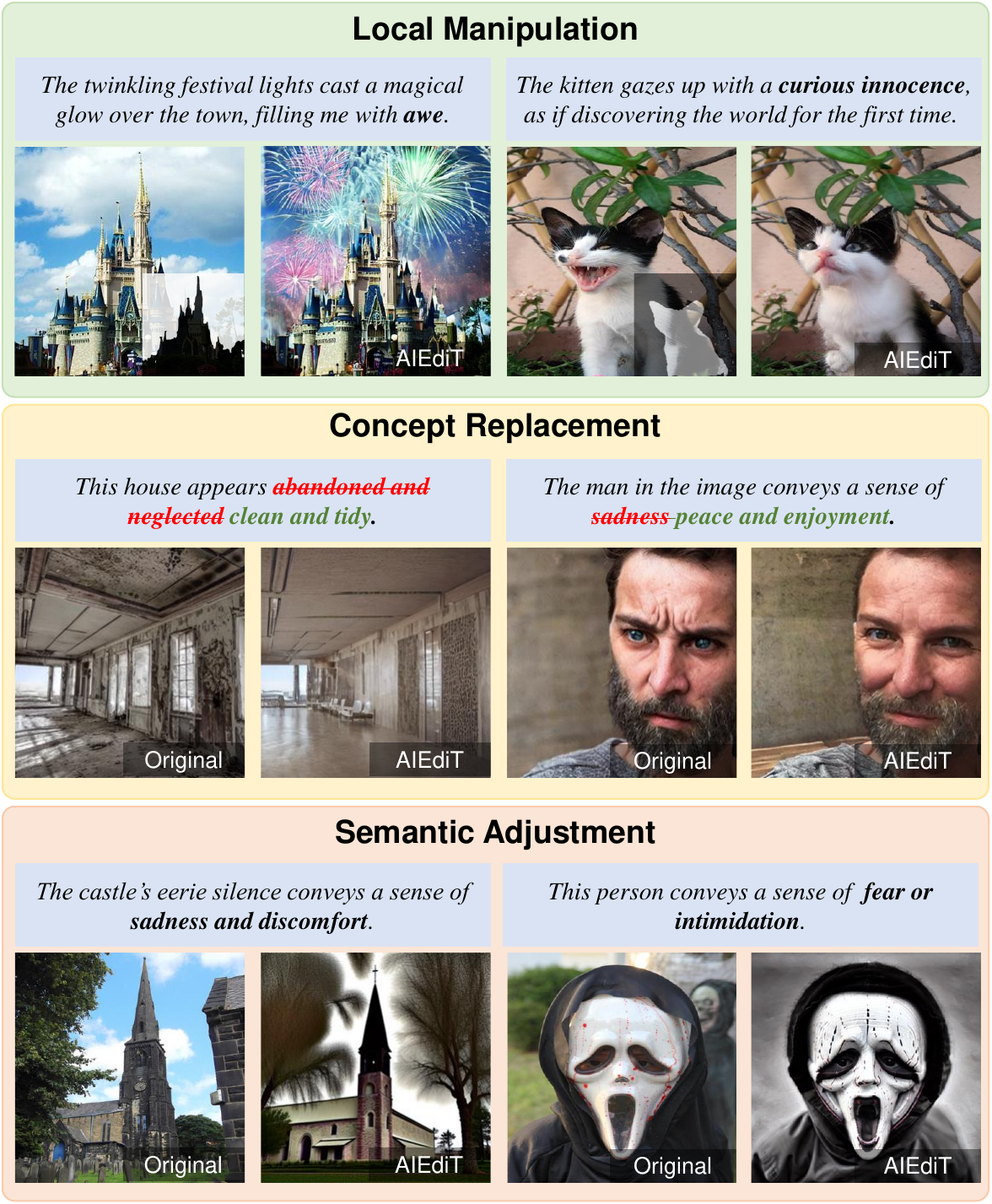}
  \caption{Additional application scenarios of AIEdiT.}
  \label{fig:application}
  \vspace{-2mm}
\end{figure}

\subsection{Application}
We present additional application scenarios of AIEdiT and show the results in \Fref{fig:application}.  Technical details of each scenario are clarified as follows:
\begin{itemize}
\item \textbf{Local manipulation.} By providing an additional mask to indicate the regions to be edited, AIEdiT is able to preserve the remaining regions unchanged. Specifically, during the inference process, the mask is applied to prevent specific regions from being distorted. This could be formulated as $z_{t} = z_{t}^\prime \odot \mathbf{M} + z^{\mathrm{in}}_{t} \odot (1-\mathbf{M})$, where $z_{t}^\prime$ and $z^{\mathrm{in}}_t$ are the latent codes refined from $z_{t+1}$ and distorted from the original images, respectively.  $\mathbf{M}$ refers to the binary mask provided by the user, with values of one indicating the regions to be edited and zeros denote the unchanged regions.

\item \textbf{Concept replacement.} Given affective images and corresponding text descriptions, AIEdiT is capable of replacing the original emotional concepts. Specifically, null-text inversion~\citep{mokady2023null} is applied to invert the original image with a corresponding text prompt into the domain of our AIEdiT. Following the users' instructions, the cross-attention maps is then modified to establish correspondences between the emotional concepts and emotional factors, using the Prompt-to-Prompt approach~\citep{prompt2prompt}. This allows the emotion to be changed while keeping the instance similar.

\item \textbf{Semantic adjustment.} Given affective images and target text descriptions, AIEdiT can adjust the overall image semantics to reflect different emotional requests. Specifically, the original semantics of the target text descriptions are first extracted using CLIP~\citep{clip}, which are then optimized to better align with the image semantics. Next, we fine-tune the backbone parameters while keeping the emotional mapper and text semantics fixed. After interpolating between the optimized and original semantics, we feed the results into our fine-tuned AIEdiT to generate the adjusted images.
\end{itemize}

\begin{table}[t]
    \centering
    \setlength\tabcolsep{2pt} 
    \caption{Task differences between our proposed AIEdiT and relevant affective image creation methods.}
    \vspace{2mm}
    \begin{adjustbox}{width={0.48\textwidth},totalheight={\textheight},keepaspectratio}
    \begin{tabular}{l | c c | c c | c c}
        \toprule
        \multirow{2}{*}{{Method}} & \multicolumn{2}{c|}{{Input modality}} & \multicolumn{2}{c|}{{Editing object}} & \multicolumn{2}{c}{{Creation mode}} \\
        \cmidrule{2-7}
        & Category & Text & Specificity & Variety & Generation & Editing  \\ \midrule
        EmoEditor & \checkmark &  &  & \checkmark &  & \checkmark \\ 
        EmoEdit & \checkmark &  &  & \checkmark &  & \checkmark \\
        C2A2 & \checkmark &  & \checkmark &  &  & \checkmark  \\ 
        AIFormer &  & \checkmark & \checkmark & & & \checkmark \\
        EmoGen & \checkmark &  &  & \checkmark & \checkmark & \\ 
        Ours (AIEdiT) &  & \checkmark  &   & \checkmark & \checkmark & \checkmark \\ \bottomrule
    \end{tabular}
    \end{adjustbox}
    \label{tab:discuss_with_other_mothods}
    \vspace{2mm}
\end{table}

\section{Discussion}
In this section, we illustrate in detail the task differences between AIEdiT and other relevant affective image creation methods (\Sref{sec:task_differences}). 
Then, we demonstrate the model's potential for generating affective images from random Gaussian noise, evaluating its performance qualitatively and quantitatively against relevant image generation methods (\Sref{sec:generationtask}). Finally, we present and analyze failure cases to identify areas for potential improvement (\Sref{sec:failure_case}).

\subsection{Task Differences} \label{sec:task_differences}
\begin{figure}[t]
  \centering
  \includegraphics[width=\linewidth]{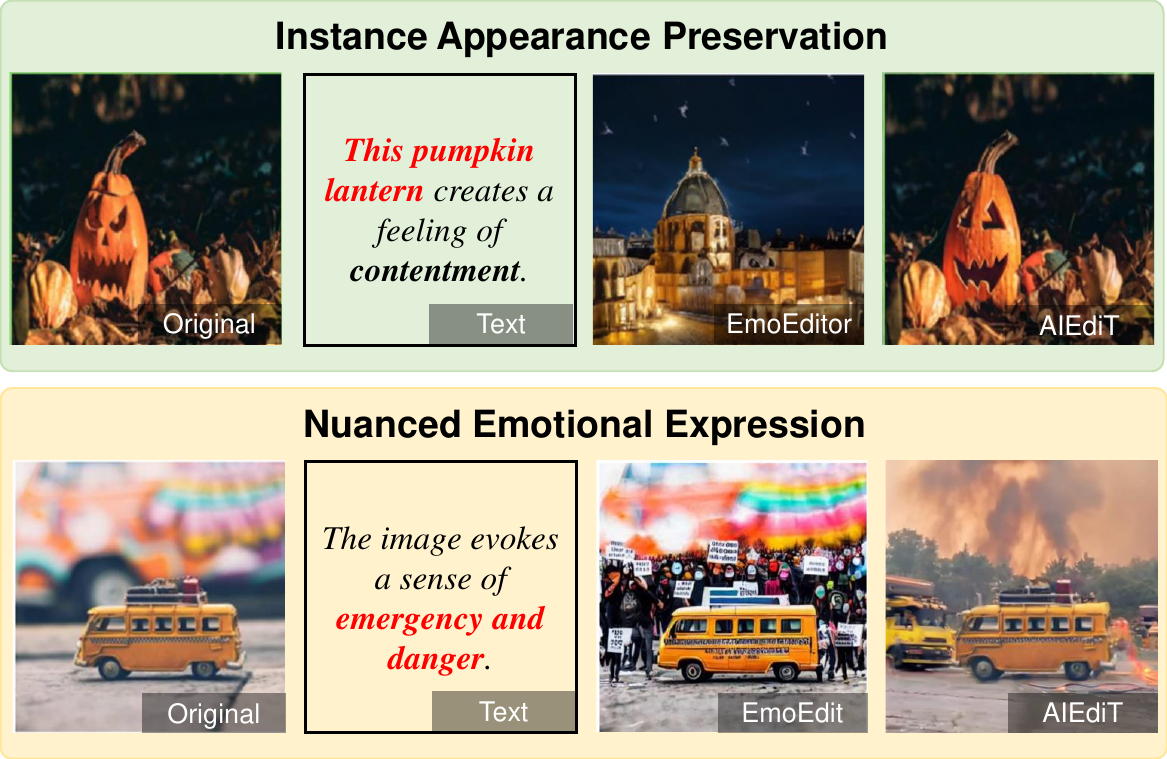}
  \caption{Illustration of our AIEdiT's advantages. Leveraging text descriptions to reshape emotional factors offers more nuanced emotional expression than emotion category-based methods.}
  \label{fig:compare_edit_supp}
\end{figure}

We summarize the differences between our proposed AIEdiT and recent affective image creation methods in~\Tref{tab:discuss_with_other_mothods}.
These methods are distinguished based on three aspects:  \textit{(i)} \textbf{Input modality:} whether the method is guided by emotion categories or text descriptions; \textit{(ii)} \textbf{Editing object:} whether the method manipulates specific properties or various factors. \textit{(iii)} \textbf{Creation mode:} whether the method functions as an image generation or editing model;
\checkmark indicates the model supports the corresponding functionality. 
We further highlight their differences as follows:
\begin{itemize}
\item Although EmoEditor~\citep{happier} and EmoEdit~\citep{emoedit} are closely related methods to edit affective images, their reliance on coarse-grained emotion categories restricts the expression of nuanced emotions and makes it challenging to preserve instance appearance during editing. We only conduct the comparison with EmoEdit~\citep{emoedit} in \Sref{sec:comparison} since EmoEditor~\citep{happier} is not publicly available. We further qualitatively compare the controllability differences in \Fref{fig:compare_edit_supp} by directly pasting their reported results.

\item C2A2~\citep{C2A2} and AIFormer~\citep{aif} focus on specific image properties (\ie, facial expression and color tune), which limits their application in general affective image editing for achieving more complex emotional editing goals. We only compare AIFormer~\citep{aif} in \Sref{sec:comparison} since C2A2~\citep{C2A2} is not publicly available.

\item EmoGEN~\citep{emogen} is an affective image generation model that produces images from scratch based on coarse-grained emotion categories, rather than editing specific input images provided by the user. We further discuss relevant generation method in \Sref{sec:generationtask}.
\end{itemize}

In contrast, leveraging the flexibility of text descriptions, AIEdiT can specify the instance to be edited while maintaining its appearance similar and allows users to express their subjective feelings that may be difficult to categorize within a specific emotion category (\eg, \Fref{fig:compare_edit_supp} second row, the fire, smoke, and chaos create a sense of emergency and danger). Consequently, to the best of our knowledge, AIEdiT is the first method to provide comprehensive manipulation of emotional factors through fine-grained text descriptions, allowing for more nuanced emotional expression.

\begin{table}[t]
\caption{Quantitative experiment results comparing with relevant image generation methods.}\label{tab:supp_comparison}
\vspace{2mm}
\centering
\setlength\tabcolsep{8pt} 
\begin{adjustbox}{width={0.48\textwidth},totalheight={\textheight},keepaspectratio}
    \begin{tabular}{l|c c c c}
    \toprule
    Methods  & FID $\downarrow$ & Sem-C $\uparrow$   & KLD $\downarrow$  & Pref (\%) $\uparrow$  \\ \midrule
    SD  & $34.41 $  & $0.671$ & $1.2953$ & $11.56$  \\ 
    SDXL & $36.86 $ & $0.681$ & $1.2118$ & $17.56$  \\ 
    RPG  & $44.17$ & $0.678$ & $1.2171$ & $15.20$  \\ 
    PixArt  & $45.96 $ & $0.677$ & $1.2583$ & $12.40$  \\ 
    EmoGen & $57.31$ & $0.698$ & $2.1436$ & $9.16$  \\ 
    Ours (AIEdiT) & $\mathbf{33.80}$ & $\mathbf{0.705}$ & $\mathbf{1.2004}$ & $\mathbf{34.12}$   \\ 
    \bottomrule
    \end{tabular}
    \end{adjustbox}
\end{table}

\begin{figure*}[t]
  \centering
  \includegraphics[width=0.9\linewidth]{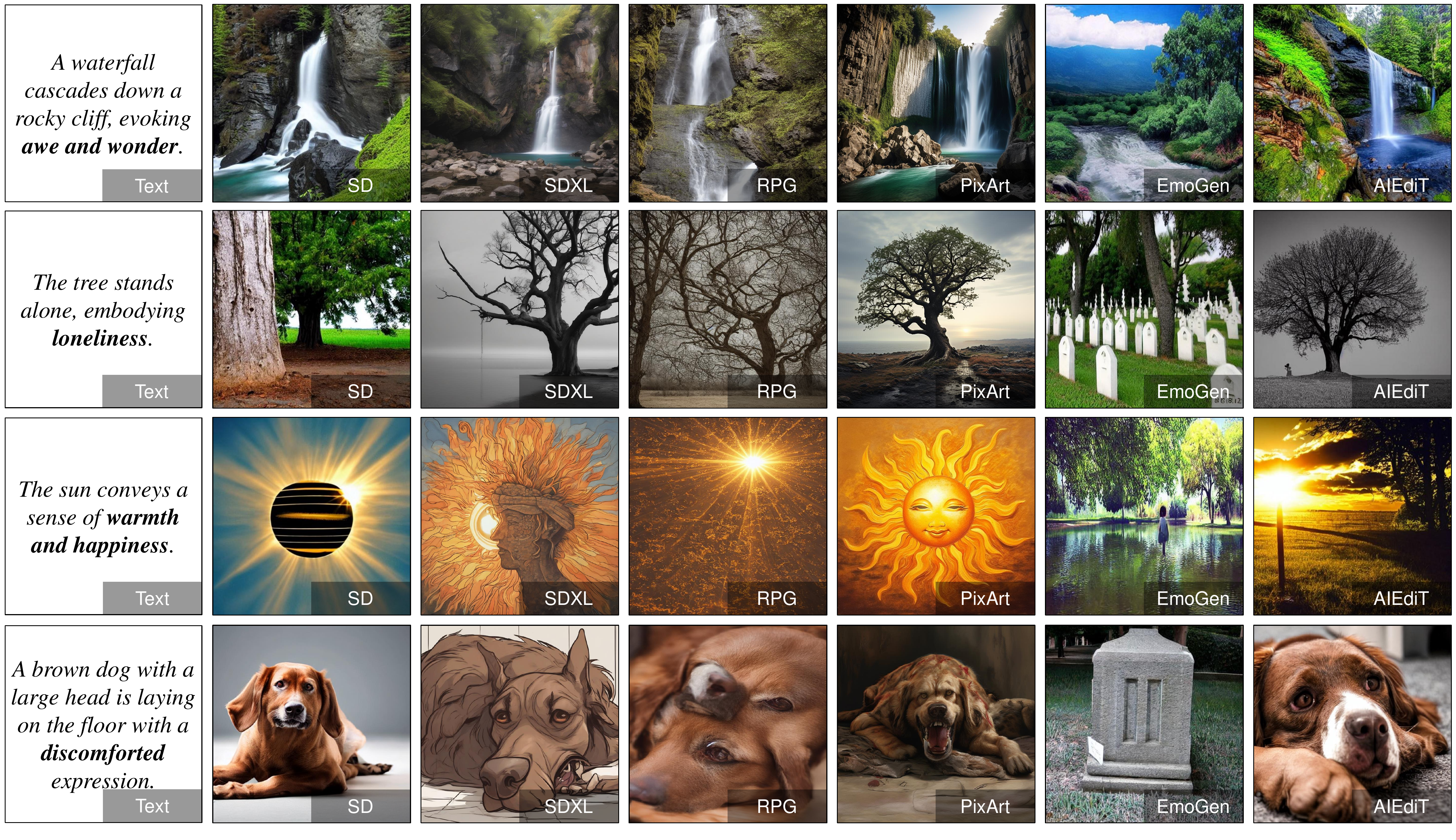}
  \caption{Visual quality comparisons with relevant image generation methods. }
  \label{fig:supp_generation_comparison}
\end{figure*}

\subsection{Performance on Image Generation Task} \label{sec:generationtask}
AIEdiT can also function as an affective image generator by editing random Gaussian noise.
To demonstrate its effectiveness in generating affective visual elements, we provide additional comparisons with relevant image generation methods, including general image generation methods (\eg, SD~\citep{stablediffusion}, SDXL~\citep{sdxl}, RPG~\citep{RPG}, PixArt~\citep{pixart}) and affective image generation method (EmoGen~\citep{emogen}).

\noindent \textbf{Qualitative comparisons.} We present the qualitative comparison results in \Fref{fig:supp_generation_comparison}. General text-driven image generation methods~\citep{pixart,sdxl,stablediffusion,RPG} struggle to translate visually-abstract emotional requests into visually-concrete semantic representations (\eg, \Fref{fig:supp_generation_comparison} third row, conveying a sense of warmth and happiness). 
On the other hand, although affective image generation method~\citep{emogen} can produce images to evoke specific emotions, it relies on coarse-grained emotion categories, still facing challenges in generating images with specific instances and nuanced emotional expressions (\eg, \Fref{fig:supp_generation_comparison} fourth row, generating a brown dog with a discomforted expression). 
Instead, our AIEdiT allows users to specify the instance and accurately shape corresponding emotional factors.

\noindent \textbf{Quantitative comparisons.} We present quantitative metrics to evaluate the performance of relevant image generation methods on the EmoTIPS dataset. As reported in~\Tref{tab:supp_comparison}, AIEdiT achieves the highest scores across all metrics, demonstrating its ability to create photo-realistic results (FID) with 
clearly recognizable visual content (Sem-C) that accurately reflect users’ emotional requests (KLD).

\begin{figure}[t]
  \centering
  \includegraphics[width=\linewidth]{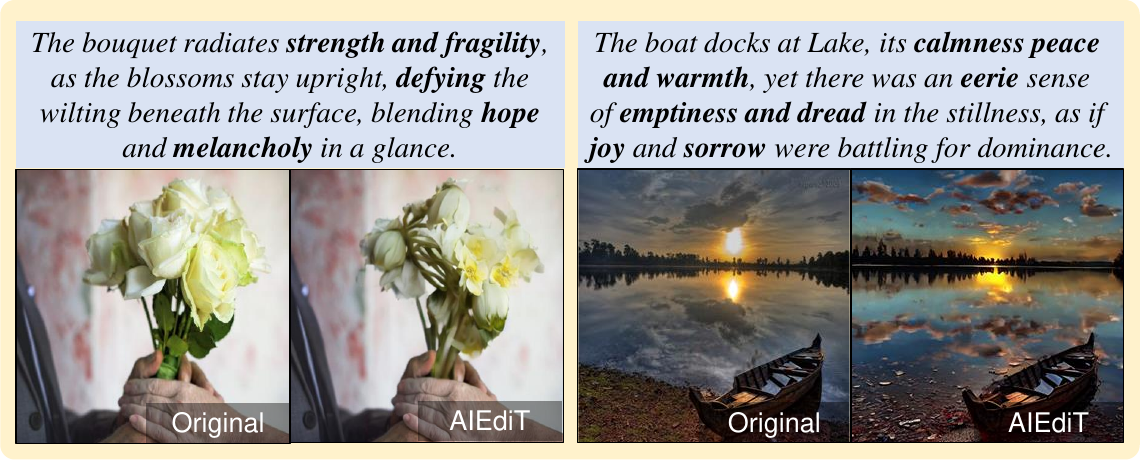}
  \caption{Visualization of failure cases. }
  \label{fig:failure_case}
\end{figure}

\subsection{Failure Case} \label{sec:failure_case}
Although AIEdiT is tailored for affective image editing, it still struggles to interpret complex text descriptions that include mixed emotions with multiple elements. To illustrate this, we present two failure cases in \Fref{fig:failure_case}. Specifically, AIEdiT has difficulty editing flowers to convey both hope and melancholy (\Fref{fig:failure_case}, left). Similarly, it struggles to effectively render both the boat and lake to express joy and sorrow (\Fref{fig:failure_case}, right).

\section{Conclusion}
In this paper, we introduce \textbf{AIEdiT}, a method for \textbf{A}ffective \textbf{I}mage \textbf{Edit}ing using \textbf{T}ext descriptions. AIEdiT introduces two key components: the emotional spectrum that represents universal emotional priors, and the emotional mapper that translates users' emotional requests into visual semantic representations. Leveraging an MLLM as a supervisor along with an appropriate sampling strategy, AIEdiT can manipulate emotional factors across the entire image according to users' instructions. Extensive experiments demonstrate the effectiveness of our approach.
Additionally, we contribute the \textbf{EmoTIPS} dataset, which provides \textbf{Emo}tion-aligned \textbf{T}ext and \textbf{I}mage \textbf{P}air \textbf{S}ets, further benefiting the community.

\noindent \textbf{Limitation.}
Our approach remains limited by the current capabilities of MLLMs in understanding subjective and ambiguous emotions. 
Although we have adopted three validation criteria to filter out unqualified samples, as shown in \Tref{tab:data_evaluation}, there are still fewer than 1.72\% failed ratings and 1.76\%-7.28\% borderline ratings. While our proposed AIEdiT demonstrates robustness in understanding nuanced emotional requests, these unqualified samples may still add noise, potentially impacting the accuracy of emotional understanding. We believe this limitation could be further alleviated with the continued advancement of MLLMs.

\vspace{-4mm}
\section*{Acknowledgement}
This work was supported in part by National Natural Science Foundation of China (Grant No. 62136001, 62088102, and U23B2052), National Science and Technology Major Project (Grant No. 2021ZD0109803), Program for Youth Innovative Research Team of BUPT (Grant No. 2023YQTD02).

\hspace*{\fill}

\vspace{-10mm}
{\small
\bibliographystyle{spbasic}
\bibliography{egbib}
}

\end{document}